\documentclass[sigconf,screen]{acmart}

\usepackage{epsfig} 
\usepackage{booktabs} 
\usepackage{multirow}
\usepackage{flushend} 
\usepackage[utf8]{inputenc}

\setcopyright{acmcopyright}
\acmPrice{15.00}
\acmDOI{10.1145/3136755.3136817}
\acmYear{2017}
\copyrightyear{2017}
\acmISBN{978-1-4503-5543-8/17/11}
\acmConference[ICMI'17]{19th ACM International Conference on Multimodal Interaction}{November 13--17, 2017}{Glasgow, UK}

\begin{document}

\title{Data Augmentation of Wearable Sensor Data for Parkinson’s Disease Monitoring using Convolutional Neural Networks}
\titlenote{This work was partly supported by the  \grantsponsor{aaa}{EU Seventh Framework Programme FP7/2007-2013} within the ERC Starting Grant Control based on Human Models (con-humo), grant agreement No. ~\grantnum{aaa}{337654}.}

\author{Terry T. Um}
\affiliation{
  \institution{University of Waterloo}
   \country{Canada}
}
\email{terry.t.um@gmail.com}

\author{Franz M. J. Pfister}
\affiliation{ 
    \institution{Ludwig-Maximilians-Univ. M{\"u}nchen}
     \country{Germany}
}
\email{fmj.pfister@me.com}

\author{Daniel Pichler}
\affiliation{%
  \institution{Technical University of Munich}
   \country{Germany}
}
\email{daniel.pichler@tum.de}

\author{Satoshi Endo, Muriel Lang, Sandra Hirche}
\affiliation{%
  \institution{Technical University of Munich}
   \country{Germany}
}
\email{{s.endo, muriel.lang, hirche}@tum.de}

\author{Urban Fietzek}
\affiliation{%
  \institution{Sch{\"o}n Klinik M{\"u}nchen Schwabing}
   \country{Germany}
}
\email{urban.fietzek@schoen-kliniken.de}

\author{Dana Kuli\'{c}}
\affiliation{%
  \institution{University of Waterloo}
   \country{Canada}
}
\email{dana.kulic@uwaterloo.ca}

\renewcommand{\shortauthors}{T. Um, F. Pfister, D. Pichler, S. Endo, M. Lang, S. Hirche, U. Fietzek, and D. Kuli\'{c}}
\renewcommand{\shorttitle}{Data Augmentation of Wearable Sensor Data for Parkinson’s Disease Monitoring ...}

\begin{abstract}
While convolutional neural networks (CNNs) have been successfully applied to many challenging classification applications, they typically require large datasets for training. When the availability of labeled data is limited, data augmentation is a critical preprocessing step for CNNs. However, data augmentation for wearable sensor data has not been deeply investigated yet.

In this paper, various data augmentation methods for wearable sensor data are proposed. The proposed methods and CNNs are applied to the classification of the motor state of Parkinson's Disease patients, which is challenging due to small dataset size, noisy labels, and large intra-class variability. Appropriate augmentation improves the classification performance from 77.54\% to 86.88\%.
\end{abstract}

%
%

\sloppy


\maketitle

\section{Introduction}

In recent years, convolutional neural networks (CNNs) have shown excellent performance on classification problems when large-scale labeled datasets are available (e.g. \cite{ResNet15, Terry16}). However, it is challenging to apply CNNs to problems where only small labelled datasets are available. For example, collecting and labeling a large amount of medical data is often difficult. As a result, it is challenging to apply CNNs to small-scale medical data.

Data augmentation leverages limited data by transforming the existing samples to create new ones. A key challenge for data augmentation is to generate new data that maintains the correct label, which typically requires domain knowledge. However, it is not obvious how to carry out label-preserving augmentation in some domains, e.g., wearable sensor data. For example, scaling of the acceleration data may change their labels because some labels are differentiated by the intensity of motion.

In this paper, the problem of classifying the motor state of Parkinson's disease (PD) patients is tackled using CNNs. PD motor state classification is a challenging task due to noisy labels, irrelevant motion interference, large variability over patients, and limited availability of the labelled data. In this paper, we propose data augmentation methods for wearable sensor data and successfully tackle the challenging PD classification task using CNNs.

The contributions of the paper can be summarized as follows:

\begin{itemize}
  \item Application of CNNs to the task of PD motor state classification, using a clinician-labeled dataset of 30 PD patients (25 patient's data are exploited) in daily-living conditions. 
  \item A set of approaches for data augmentation of wearable sensor datasets for CNN-based classification.
  \item Experimental comparison of proposed data augmentation methods.
\end{itemize}

\section{Related work} \label{Sec_RelWorks}

Most PD patients experience motor fluctuations, which are characterized by phases of bradykinesia, i.e. underscaled and slow movement, and dyskinesia, i.e. overflowing spontaneous movement \cite{PD12}. Dopaminergic treatment can alleviate symptoms of bradykinesia while its over-treatment can cause dyskinesia. Thus, an accurate evaluation of a patient's phenomenology is needed for determining the right dose of medication. Current PD motor state evaluation methods rely on patient self-reports and visual observation by the clinician \cite{PD12}.

Researchers have proposed automating the evaluation with wearable sensors (e.g. \cite{PDWearable09, PDDL16}). However, most approaches to date have been limited to standardized motor tasks in clinical settings \cite{PDReview13}. To enable automated evaluation of PD motor states which covers a wide range of PD symptoms across patients, a large amount of wearable sensor data in daily-living conditions is needed \cite{PDFreeLiving16}. Deep learning (DL) approaches \cite{DLNature15} provide a promising methodology to deal with the large variability of PD data \cite{PDLargeScale16, PDDL15, PDDL16}. Given the difficulty in collecting such large datasets, data augmentation is needed \cite{DLBook16}.

Data augmentation is an indispensable preprocessing step for achieving peak performance in DL approaches (e.g. \cite{AlexNet12, ResNet15}). For augmenting time-series data, Le Guennec et al. \cite{DAWearable16} used window slicing and window warping methods, which extracts multiple small-size windows from a single window and lengthens/shortens a part of the window data, respectively. Unlike data augmentations for image \cite{DAImage14} and speech recognition \cite{DASpeech15}, however, data augmentation for wearable sensor data has not been systematically investigated yet to the best of our knowledge. In this paper, we propose various data augmentation methods that enable the classification of PD motor states from wearable data and evaluate them using CNN. 


\section{PD Motor State Classification} \label{Sec_PDClass}

\subsection{Challenges in PD Data} \label{Sec_Challenges}

\begin{figure}[t]
    \centering
    \includegraphics[trim={0cm 0cm 0cm 0cm},width=0.4\textwidth]{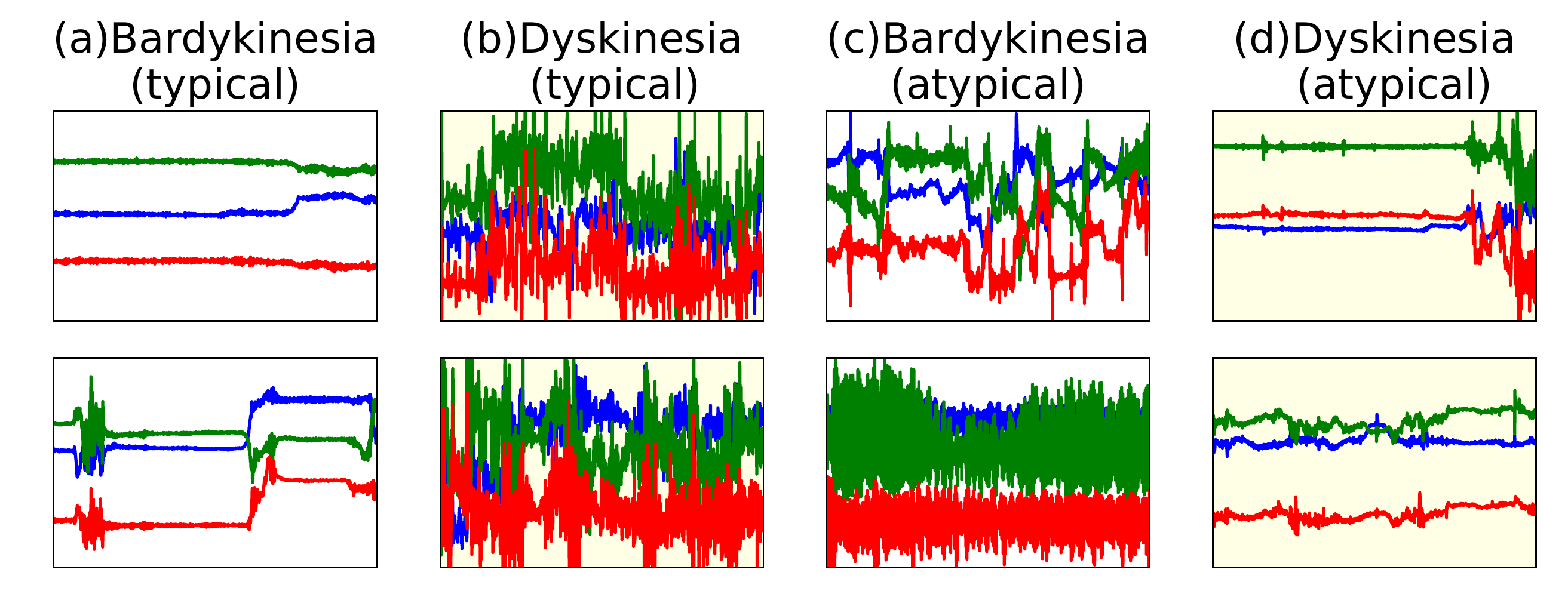}
    \caption{(a) and (b) show typical examples of bradykinesia and dyskinesia in a 1 min window while (c) and (d) show atypical patterns. The blue, red, green represent X,Y,Z signals from the accelerometer, respectively.}
    \label{fig:ex}
\end{figure}

We consider two frequent PD motor states: bradykinesia, which is characterized by decreased movement speed and may be accompanied by tremor, and dyskinesia, which is characterized by involuntary extremity movements. Figure \ref{fig:ex} illustrates exemplar one minute data windows of both motor states, from a single accelerometer worn on the wrist of PD patients. Bradykinesia data typically appear as constant signals indicating less movement (Fig \ref{fig:ex}(a)) while dyskinesia data consist of fluctuating movements (Fig \ref{fig:ex}(b)). 

However, there are a significant number of examples that deviate from the stereotypical expressions. For example, bradykinesia accompanied by tremor can show fluctuating signals which look like a dyskinesia state (Fig \ref{fig:ex}(c)). On the other hand, dyskinesia with voluntary suppression can show constant signals which look like a bradykinesia state (Fig \ref{fig:ex}(d)).

There are several factors that can cause an apparent disagreement between the observed data pattern and the expert label. First, if the body of the patient indicates, e.g., a dyskinesia state, but the hand which wears the wearable sensor does not move because the patient is, e.g., holding a chair for suppressing the symptom, the assigned label based on the overall body expression will be mismatched with the recorded data from the wearable device. Also,
the expert rater typically rates the symptoms for a fixed length window, but arbitrary segmentation into fixed length windows may not result in single motor state windows. Furthermore, the interference of voluntary movements, e.g., waving the hand, can make bradykinesia states look like dyskinesia, and, e.g, voluntary rest, appear like bradykinesia. Finally, bradykinesia accompanied by tremor can also can make it difficult to distinguish between bradykinesia and dyskinesia.

The factors described above introduce noisy labels, and lead to large intra-class variability and significant overlap between two classes. As a result, it makes the PD motor state classification more challenging, particularly given a small amount of data.

\begin{figure*}[t]
    \centering
    \includegraphics[trim={0.7cm 0.4cm 0.7cm 0cm}, width=0.95\textwidth]{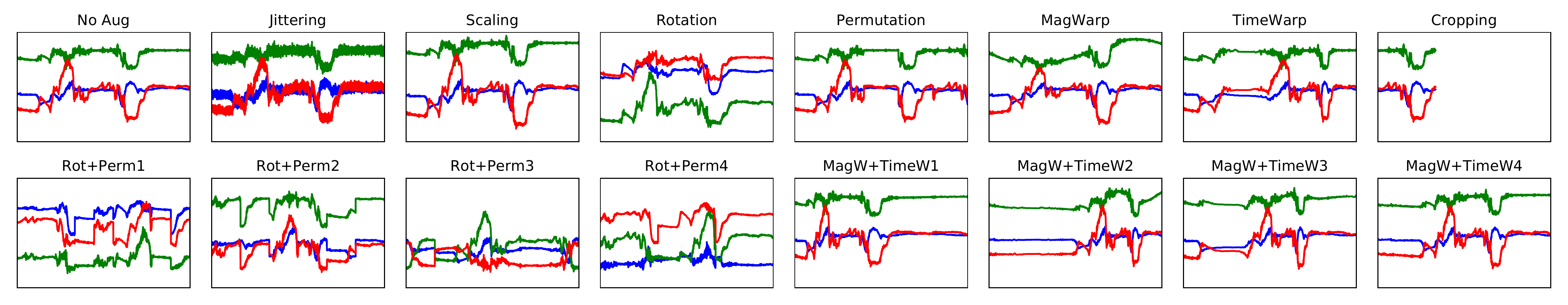}
    \caption{Various data augmentations that are used in the experiments: jittering, scaling, rotating, permutating, magnitude-warping, time-warping methods. Combinations of various data augmentations can also be applied.}
    \label{fig:aug}
\end{figure*}

\subsection{Data Augmentation Methods for Wearable Sensor Data} \label{Sec_DA}

Data augmentation can be viewed as an injection of prior knowledge about the invariant properties of the data against certain transformations. Augmented data can cover unexplored input space, prevent overfitting, and improve the generalization ability of a DL model \cite{DLBook16}. In image recognition, it is well-known that minor changes due to jittering, scaling, cropping, warping and rotating do not alter the data labels because they are likely to happen in real world observations. However, label-preserving transformations for wearable sensor data are not obvious and intuitively recognizable (Fig \ref{fig:aug}).


One factor that can introduce label-invariant variability of wearable sensor data are differences in sensor placement between participants. For example, an upside-down placement of the sensor can invert the sign of the sensor readings without changing the labels. Therefore, augmentation by applying arbitrary \textbf{rotations (Rot)} to the existing data can be used as a way of simulating different sensor placements.


Another factor that can introduce variability is the temporal location of activity events, e.g., tremor, in the window. Since the fixed size window segmentation is arbitrary, the location of the observed symptom in the window does not have any meaning. Thus, we may augment data by perturbing the location of the windows or events.


\textbf{Permutation (Perm)} is a simple way to randomly perturb the temporal location of within-window events. To perturb the location of the data in a single window, we first slice the data into $N$ same-length segments, with N ranging from 1 to 5, and randomly permute the segments to create a new window. \textbf{Time-warping (TimeW)} is another way to perturb the temporal location. By smoothly distorting the time intervals between samples, the temporal locations of the samples can be changed using time-warping.

Small changes in magnitude may preserve the labels, depending on the target task. \textbf{Scaling (Scale)} changes the magnitude of the data in a window by multiplying by a random scalar, while \textbf{magnitude-warping (MagW)} changes the magnitude of each sample by convolving the data window with a smooth curve varying around one. In addition, \textbf{jittering (Jitter)} is also considered as a way of simulating additive sensor noise. These data augmentation methods may increase robustness against multiplicative and additive noise and improve performance.

Lastly, \textbf{cropping (Crop)}, which is similar to image cropping or window slicing in \cite{DAWearable16}, is applied for diminishing the dependency on event locations. Note that cropping can capture an event-free region, which might change the label. Also, note that cropping with random locations over epochs will eventually converge to a sliding window method with arbitrary stride sizes.

In a nutshell, jittering, scaling, cropping, rotating, permutating, magnitude-warping and time-warping methods are applied for augmenting wearable sensor data. In the next section, the performance of PD motor state classification with the proposed data augmentation methods is evaluated using CNNs.

\section{Experiments} \label{Sec_Experiment}
\subsection{Data Preparation} \label{Sec_Data}

A dataset of 30 patients' motor states was collected using Microsoft Band 2 \cite{MSBand2} in daily-living conditions without requesting specific motor tasks\footnote{The study was approved by
the ethics committee of Technical University of Munich (Az. 234/16 S).}. The 30 PD patients are $67\pm10$ years old, median Hoehn \& Yahr stage $2$, average disease duration $11\pm5$ years, and MoCA points $26\pm3$. Among them, 25 patient's data are used for this research and each one minute interval is labeled by a clinical expert. The data are collected at a frequency of 62.5Hz and resampled to 120Hz to deal with sampling irregularities. The first 58-seconds of data (6960 samples) from each one minute window is used to make same-length instances.

Similar to previous works (e.g. \cite{PDWearable09}, \cite{PDHome12}, \cite{PDDL15}, \cite{PDDL16}) acceleration data only are used for the PD motor state classification. Also, \emph{no-symptom} data are removed to simplify the problem and focus on characterizing data augmentation methods. Data collected during walking, laying and eating activities are also removed due to limited observation of movement during these activities. Note that no other preprocessing, e.g., data normalization or smoothing, is applied because they may confound the data label and subsequent results.

The resulting dataset consists of 3530 min (58.8 hours) of bradykinesia and dyskinesia data. For cross-validation, the 25 PD patients are divided into five subject groups. The performance of PD motor state classification is reported in Section \ref{Sec_Result} using the average values of 5-fold cross-validation results.

\subsection{The CNN Architecture} \label{Sec_CNN}

\begin{figure}[]
	\centering
    \includegraphics[width=0.45\textwidth]{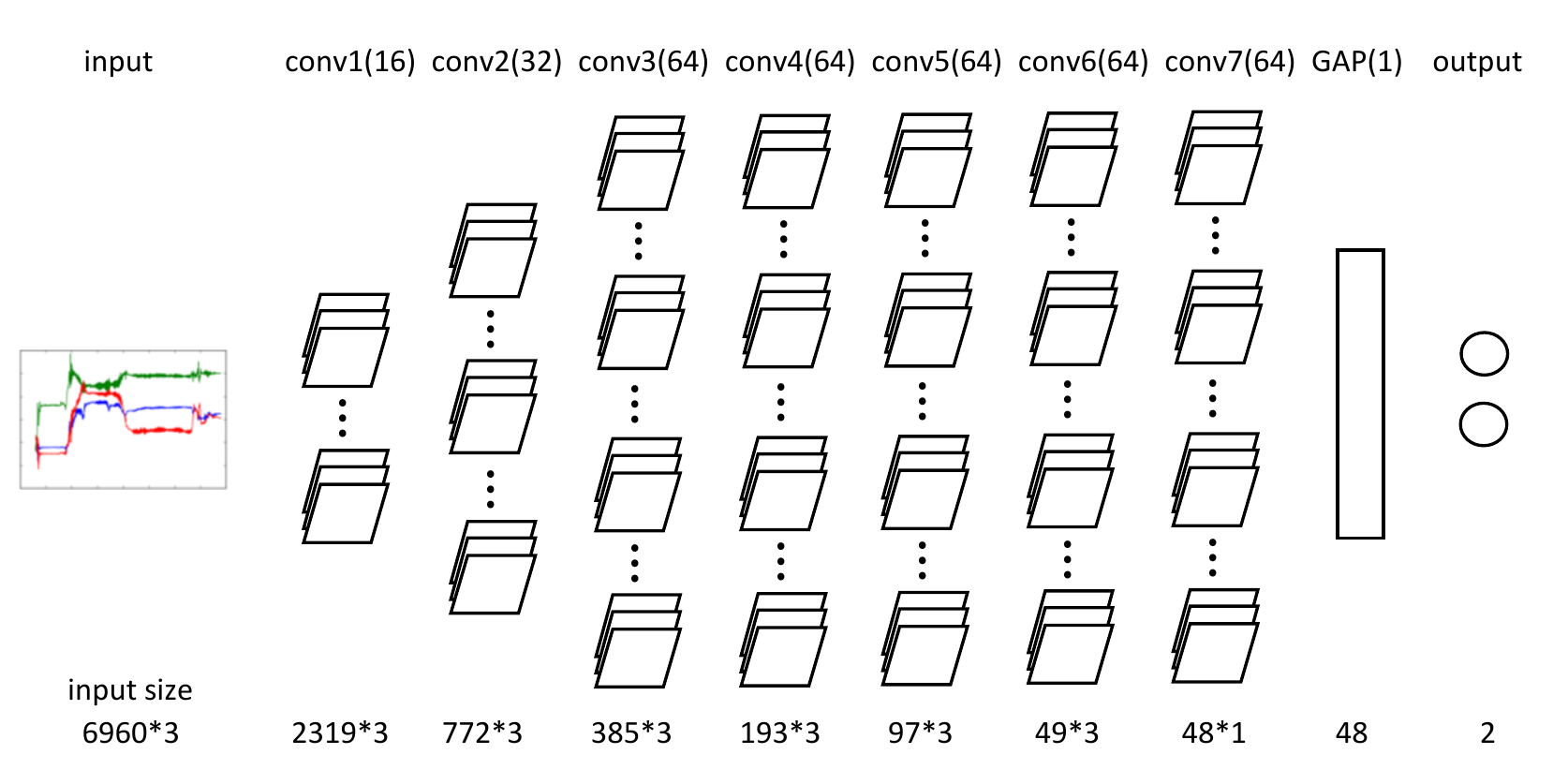}
    \caption{7-layer CNN with a global average pooling (GAP). The 7-layer CNN consists of 16-32-64-64-64-64-64 feature maps which reduce the size of the inputs to 2319*3, 772*3, 385*3, 193*3, 97*3, 49*3, 48*1, respectively.}
    \label{fig:cnn}
\end{figure}

In this research, CNNs are used for PD motor state classification. CNNs are more suitable for small-scale datasets than long short-term memories (LSTMs) \cite{LSTM1997} because CNNs generally use a smaller number of parameters compared to fully-connected LSTMs. Deep and sparse 7-layer CNNs (Figure \ref{fig:cnn}) are employed to capture the large variability of the small-scale PD data.

A convolutional layer, a batch normalization layer \cite{BatchNorm15}, and an activation layer using rectified units (ReLUs) form a single convolutional layer of the 7-layer CNN. With strided convolutions using 4*1, 4*1, 3*1, 3*3, 2*3, 2*3, 2*3 convolution filters, the sizes of the inputs are reduced from 6960*3 to 48*1 over layers (Figure \ref{fig:cnn}). Note that XYZ signals of the accelerometer are convolved in layers 4,5,6 and 7 to capture inter-vector-component features. For reducing the number of parameters for small-scale datasets, a global averaging pooling (GAP) layer \cite{NIN13} is applied at the end instead of fully-connected layers.

\subsection{Results} \label{Sec_Result}

Classification of PD motor states is performed using the CNN with the various data augmentation methods. For baseline results, a support vector machine (SVM) with an RBF kernel is applied to 540 dimensional statistical features: mean, variance, skewness, kurtosis, and maximum values are extracted from 1 min data using 5 and 10-sec sliding windows. Also, a CNN is applied to raw 1 min data without data augmentation for baseline comparison. All experiments except for the SVM are performed for 400 epochs and the median values from the last 10 epoch results are used for averaging the 5-fold cross-validation results.

Different random parameter values are applied for data augmentation. For jittering, a standard deviation (STD) value is sampled from a Gaussian distribution with 0.03 STD, and 1 min of Gaussian noise is generated using the sampled STD value. For scaling, a random scalar is sampled from a Gaussian distribution with a mean of 1 and 0.1 STD. For rotation, an arbitrary rotation matrix is generated for each instance. For permutation, a random integer $N$ is determined by rounding a positive value sampled from a Gaussian distribution with 5.0 STD. For magnitude-warping and time-warping, random sinusoidal curves are generated using arbitrary amplitude, frequency, and phase values. The implemented code for the proposed data augmentation methods is available online: \url{https://github.com/terryum/Data-Augmentation-For-Wearable-Sensor-Data}

\begin{table}[]
\centering
\caption{The results of PD motor state classification with various data augmentation methods. R,P,T,M represent \emph{Rot}, \emph{Perm}, \emph{TimeW}, \emph{MagW}, respectively.}
\label{table:result}
\begin{tabular}{@{}cccccccc@{}}
\toprule
      & SVM   & CNN   & Jitter & Scale & Crop  & Rot   & Perm    \\ \midrule
Train & 98.82 & 99.92 & 99.78  & 99.84 & 65.77 & 100.0 & 99.33   \\
Test  & 70.72 & 77.54 & 77.52  & 79.46 & 73.58 & \textbf{82.62} & 81.16   \\ \toprule
      & MagW  & TimeW & P,T    & R,P   & R,T   & R,P,T & R,P,T,M \\ \midrule
Train & 100.0 & 94.67 & 96.63  & 99.08 & 94.70 & 94.43 & 94.20   \\
Test  & 79.33 & 82.00 & 81.75  & \textbf{86.76} & 85.01 & \textbf{86.88} & 85.60   \\ \bottomrule
\end{tabular}
\end{table}

\begin{figure}[t]
    \centering
    \includegraphics[width=0.45\textwidth]{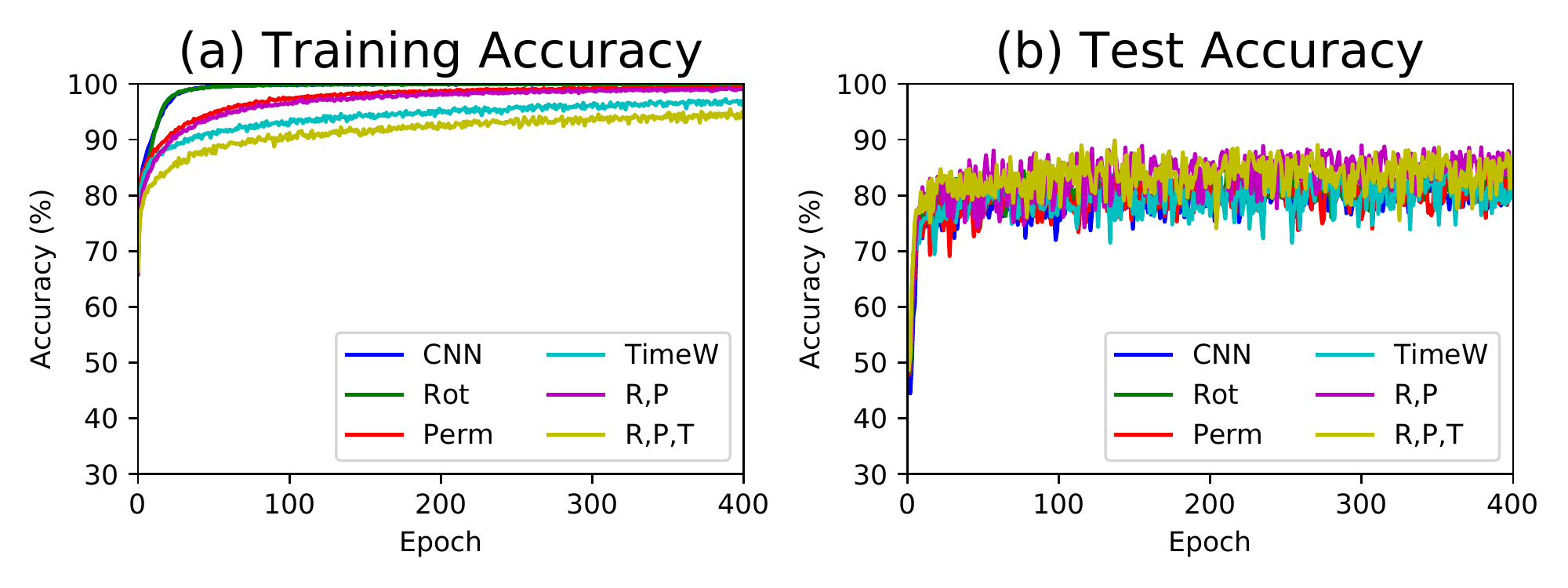}
    \caption{Training curves for \emph{CNN}, \emph{Rot}, \emph{Perm}, \emph{TimeW}, \emph{Rot+Perm} and \emph{Rot+Perm+TimeW} methods. The curves of \emph{Rot+Perm+TimeW} shows slow training improvement and a better generalization performance.}
    \label{fig:train}
\end{figure}

The main results are presented in Table \ref{table:result}. Jittering fails to improve the performance of PD motor state classification because it introduces rapid fluctuations which look similar to dyskinesia. Cropping also fails because it drops the information of $2/3$ window samples, which could be a critical loss given the small dataset. Cropping of an event-free region also hinders the learning process and can be a cause of the poor performance. Scaling and magnitude-warping also fail because changing of the intensity of the signal may alter the labels.


On the other hand, rotation, permutation, and time-warping methods improve the performance of PD motor state classification. The best performance among the single data augmentation methods is achieved by rotation. Permutation and time-warping also provide performance improvements by perturbing the temporal locations of samples. These results indicate that the major sources of variability are different sensor placements between participants and event locations in an arbitrarily segmented window. The proposed rotation, permutation, and time-warping methods effectively compensate the unnecessary variations and improve the performance by 3.6-5.1\% accuracy.





Combinations of various data augmentation methods show better performance than that of a single data augmentation method. The combinations of \emph{Rot}+\emph{Perm} and \emph{Rot}+\emph{TimeW} show better performance than the baseline of \emph{CNN} by 7.5-9.2\%. The best performance is achieved by \emph{Rot}+\emph{Perm}+\emph{TimeW} with 86.88\% accuracy. These results indicate that rotation can be used to alleviate sensor pose variability while either permutation or time-warping can be employed for addressing the variability of temporal locations of events in a window.

\begin{figure}[t]
    \centering
    \includegraphics[width=0.48\textwidth]{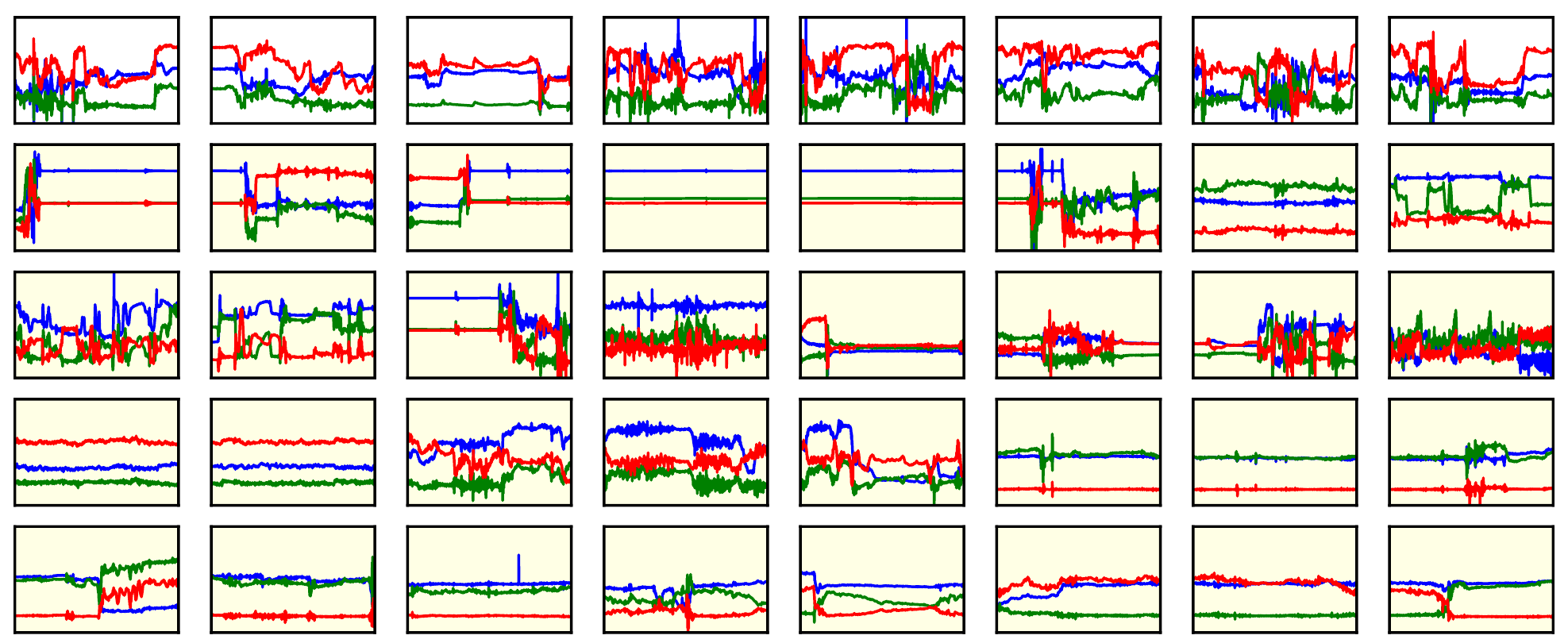}
    \caption{Randomly selected 40 incorrect mispredictions from the Fold-1 results of \emph{Rot}+\emph{Perm}+\emph{TimeW} experiment. Fluctuating signals from bradykinesia (white) and constant signals from dyskinesia (yellow) are often misclassified.}
    \label{fig:wrong}
\end{figure}

Training curves of the experiments are depicted in Fig \ref{fig:train}. The \emph{Rot}+\emph{Perm}+\emph{TimeW} curve shows slow training improvement and a better generalization performance than others thanks to the regularization effect provided by the data augmentation. Some of the failed predictions are presented in Fig \ref{fig:wrong}. From the figure, it can be observed that CNNs often misclassify fluctuating bradykinesia and constant dyskinesia data, which can be considered as seemingly-noisy labels as described in Section \ref{Sec_Challenges}.





\section{Conclusion} \label{Sec_Conclusion}
In this paper, an automatic classification algorithm for PD motor state monitoring is developed based on wearable sensor data. PD motor state classification is a challenging task because of large inter-class variability, noisy labels, interference by irrelevant motion signals and limited data availability. The challenging PD task is successfully tackled using a 7-layer CNN and the proposed data augmentation methods. The combination of rotational and permutational data augmentation methods improves the baseline performance of 77.52\% accuracy to 86.88\%. Systematic experiments with various data augmentation methods provide a direction towards a general approach for augmentation for wearable sensor data.




\newpage
\bibliographystyle{ACM-Reference-Format}
\bibliography{PD2C_ICMI2017}

\end{document}